%% file: acl_latex.tex
\pdfoutput=1

\documentclass[11pt]{article}
\usepackage[dvipsnames, svgnames, x11names]{xcolor}

\usepackage[]{acl}

\usepackage{times}
\usepackage{latexsym}

\usepackage{CJKutf8}
\usepackage{booktabs}
\usepackage{graphicx}
\usepackage{multirow}
\usepackage{makecell}
\usepackage{amssymb}
\usepackage{pifont}
\usepackage{arydshln}
\usepackage{natbib}
\usepackage{color}
\usepackage{wasysym}
\usepackage{subcaption}
\usepackage{comment}
\usepackage{verbatim}
\usepackage{fancyvrb}
\usepackage{xcolor}
\usepackage{multicol}
\usepackage{tcolorbox}
\usepackage{inconsolata}
\usepackage{float}
\usepackage{url}

\usepackage[T1]{fontenc}

\usepackage[utf8]{inputenc}

\usepackage{microtype}

\usepackage{inconsolata}

\usepackage{graphicx}

\title{How Far Can In-Context Alignment Go?\\Exploring the State of In-Context Alignment}

\author{Heyan Huang$^{12}$~~~Yinghao Li$^{1}$~~~Huashan Sun$^{1}$~~~Yu Bai$^{1}$~~~Yang Gao$^{12}$ \\
$^{1}$School of Computer Science and Technology, Beijing Institute of Technology, Beijing, China \\
$^{2}$Beijing Institute of Technology Southeast Academy of Information Technology, Putian, China \\
\texttt{\{hhy63,yhli,hssun,yubai,gyang\}@bit.edu.cn}
}

\begin{document}
\begin{CJK}{UTF8}{gbsn}

\maketitle
\begin{abstract}

Recent studies have demonstrated that In-Context Learning (ICL), through the use of specific demonstrations, can align Large Language Models (LLMs) with human preferences known as In-Context Alignment (ICA), indicating that models can comprehend human instructions without requiring parameter adjustments. However, the exploration of the mechanism and applicability of ICA remains limited.
In this paper, we begin by dividing the context text used in ICA into three categories: format, system prompt, and example. Through ablation experiments, we investigate the effectiveness of each part in enabling ICA to function effectively. We then examine how variants in these parts impact the model's alignment performance. Our findings indicate that the example part is crucial for enhancing the model's alignment capabilities, with changes in examples significantly affecting alignment performance.
We also conduct a comprehensive evaluation of ICA's zero-shot capabilities in various alignment tasks. The results indicate that compared to parameter fine-tuning methods, ICA demonstrates superior performance in knowledge-based tasks and tool-use tasks. However, it still exhibits certain limitations in areas such as multi-turn dialogues and instruction following.

\end{abstract}

\section{Introduction}
\label{intro}

Pre-trained large language models (LLMs) typically require an instruction fine-tuning phase during which model parameters are adjusted to align the model with human preferences, enabling it to follow human instructions and function as an interactive assistant~\citep{Wei2021FinetunedLM,NEURIPS2022_b1efde53,badola-etal-2023-parameter,ghosh2024closer}. 
Although this phase is much less costly than pre-training, it still involves complex and challenging issues, such as problems related to data mixture ratio and quality, as well as the problem of knowledge forgetting~\citep{NEURIPS2023_ec641387,dong2024abilities}.

\begin{figure}[t]
    \centering
    \includegraphics[width=0.95\linewidth]{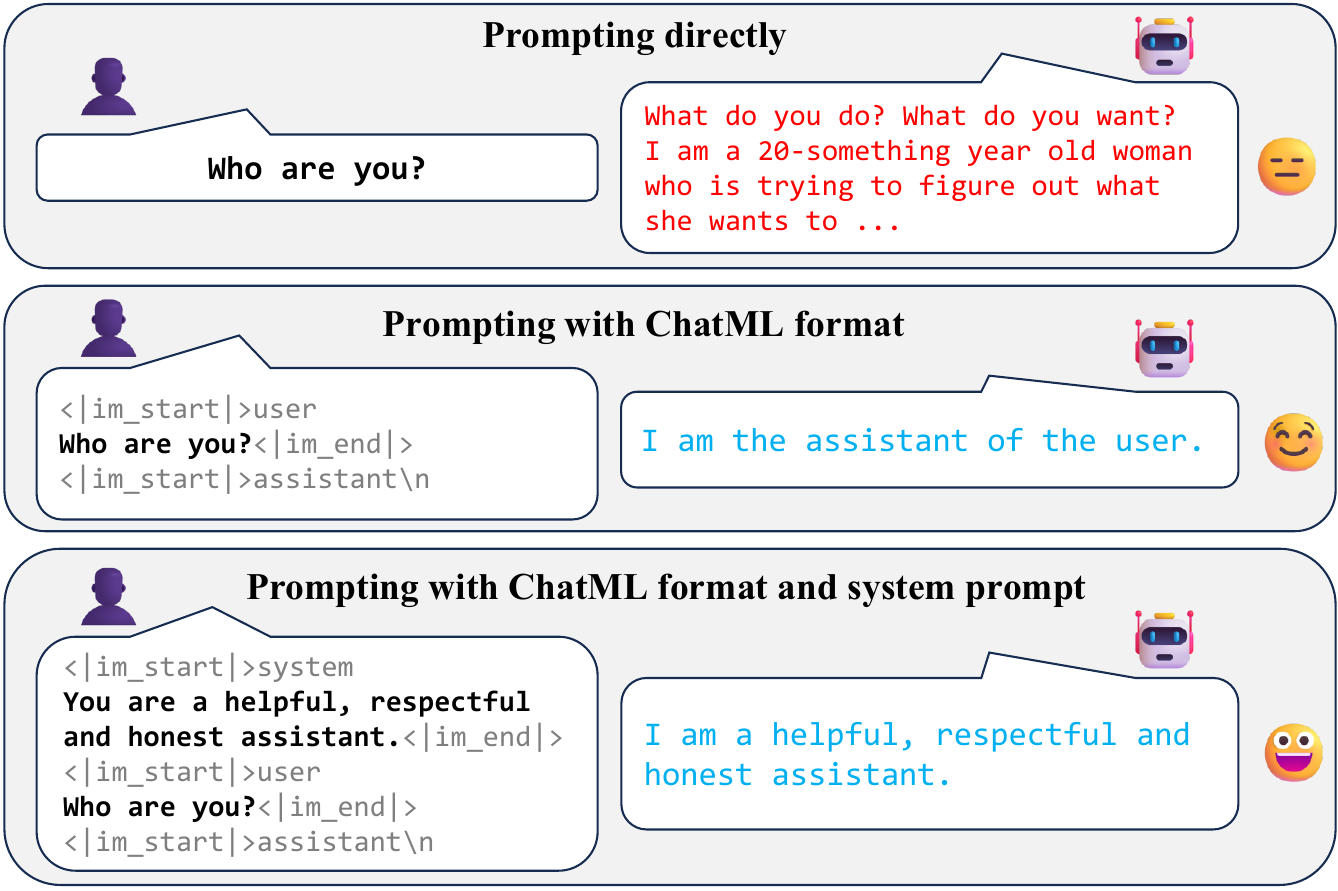}
    \caption{ 
    Responses from Llama2-7b-base model with different prompting styles. Directly prompting the model results in repetitive and meaningless responses. With ChatML format, the model understands and responds appropriately to the user's query. Further adding a system prompt makes the model's responses more comprehensive. (\textit{The model does not automatically stop generating but is truncated with ``<|im\_end|>''.})
    }
    \label{fig:prompt}
\end{figure}

Recent studies have proposed that by crafting specific demonstrations, base models can achieve alignment with human preferences through In-Context Learning (ICL). This process is known as In-Context Alignment (ICA)~\citep{Lin2023ReAlign, han2023context}. 
ICA enables models to understand instructions without the need for parameter fine-tuning. 
This cost-effective nature positions ICA as a potentially viable alternative to the fine-tuning-based alignment methods. 
However, the exploration of ICA in these studies has been limited to exploring the alignment capabilities of LLMs as open-domain assistants in single-turn dialogue. 
Detailed analysis lacks regarding the underlying mechanisms and broader applicability of ICA, leading to our main research question: How far can ICA go?

This raises several questions: 1. Does ICA rely on specific parts or the entirety of the context to align with human preferences? 2. Considering the significance of various contextual parts, would substituting them with alternative variants affect ICA performance to varying extents? 3. With a preferable setting identified by exploring the previous question, what other tasks can ICA effectively perform beyond single-turn dialogue? These questions are related to each other and we propose to address them sequentially.

To begin exploring the first question, we initiate a pilot ICA experiment with a pre-trained model, as illustrated in Figure~\ref{fig:prompt}. The purpose is to investigate the alignment effect of format and system prompt in the simplest scenario.
We apply the ChatML\footnote{\url{https://github.com/MicrosoftDocs/azure-docs/blob/main/articles/ai-services/openai/how-to/chat-markup-language.md}} format, commonly used for fine-tuning models, to Llama2-7b-base model~\citep{Touvron2023Llama2O}. To our surprise, the model can understand user queries and generate meaningful responses, a capability that direct prompting methods could not achieve.
By probing the effects of the format and system prompt, we further explore the roles that these parts, along with the examples, play in ICA.

In this paper, we initially explore how contextual content affects ICA by dividing it into three parts: format, system prompt, and example (Section~\ref{what}).
Building on this, we address the second question by designing variants for each part to investigate their impact on ICA performance (Section~\ref{which}).
Finally, for the third question, we explore the effectiveness of ICA in other instruction tasks (Section~\ref{how}). Additionally, we train two SFT models and compare them with the original chat model. 
Our contributions can be summarized as follows:

\begin{enumerate}
    \item We explore the impact of format, system prompt, and example on In-Context Alignment (ICA), demonstrating that the example part is the most crucial for ICA.

    \item We show that different variants of the example have varying impacts on ICA, and that ICA based on large-parameter models can surpass fine-tuned models of the same scale.

    \item We demonstrate that ICA outperforms fine-tuned models in knowledge-based and tool utilization tasks, although there is still a gap in performance for multi-turn dialogue and instruction-following tasks.

\end{enumerate}

\section{Related Work}

\subsection{Fine-tuning Based Alignment}

Fine-tuning-based alignment refers to the process of adjusting model parameters to produce responses that align with human preferences.

\paragraph{Supervised Fine-Tuning (SFT)}
Currently, SFT has emerged as the most prevalent method for achieving alignment~\citep{10057013,lu2023instag,PAN2023198}. Given the high costs associated with acquiring high-quality, manually crafted instruction datasets, the Self-instruct method~\citep{wang2023selfinstruct} has gained popularity for automatically generating instruction data from large models. For instance, Alpaca~\citep{alpaca} employs the self-instruct approach to generate a dataset of 52,000 instructions for training. Fine-tuning the 7B parameter Llama model requires about three hours of training on eight 80GB A100 GPUs. Despite the relatively short training duration, full parameter fine-tuning demands substantial hardware resources.

\paragraph{Parameter-Efficient Fine-Tuning (PEFT)}
PEFT ~\citep{peft} methods only require fine-tuning a small number of model parameters to efficiently adapt large pre-trained models to various downstream applications~\citep{hu2021lora,frantar2023gptq,NEURIPS2023_1feb8787}.
The LoRA method~\citep{hu2021lora} addresses this by injecting trainable low-rank decomposition matrices into each layer of the Transformer architecture~\citep{vaswani2017attention}, thereby significantly reducing the number of trainable parameters required for downstream tasks and expediting the SFT training process. Furthermore, introducing the ``Superficial Alignment Hypothesis'', \citet{NEURIPS2023_ac662d74lima} demonstrated that a dataset of just 1000 high-quality, manually written instructions (LIMA) could achieve effective alignment, laying a foundational hypothesis for the feasibility of ICA.

Fine-tuning-based alignment methods still require certain training resources and face some difficult-to-solve issues~\citep{gudibande2023false,gekhman2024does}, which makes ICA more valuable for research.

\subsection{In-Context Alignment (ICA)}

ICA refers to using in-context learning with carefully designed prompts to achieve alignment without adjusting the model's parameters.

\paragraph{In-Context Learning (ICL)}
Since the discovery of ICL capabilities in LLMs~\citep{NEURIPS2020_1457c0d6}, there has been a growing body of research exploring the underlying mechanism and applications of ICL~\citep{bai2024identifying,pmlr-v237-abernethy24a}. For instance, \citet{von2023transformers} and \citet{dai-etal-2023-gpt} examined the mechanism of ICL from the perspective of gradient descent learning, suggesting that ICL functions as an implicit fine-tuning method. Other studies have investigated how contextual examples impact model performance. \citet{min2022rethinking} demonstrated that randomly replacing labels in contextual demonstrations has minimal effect on the performance of various classification and multiple-choice tasks. \citet{wu-etal-2023-self} introduced a self-adaptation mechanism for selecting and arranging contextual examples, thereby improving the model's few-shot learning capabilities. \citet{li-qiu-2023-finding} proposed a metric to assist the model in determining the optimal arrangement of examples.

\paragraph{Alignment With ICL}
Earlier research on ICL mainly focused on tasks such as classification and multiple-choice questions. However, recent work has started to explore the application of ICL to a wider array of tasks. \citet{Ye2023InvestigatingTE} explored the direct relationship between ICL and instructions, demonstrating that inserting task-irrelevant prompts in the input can also enhance the instruction-following capabilities of large language models (LLMs) during reasoning. \citet{han2023context} applied ICL to open-domain dialogue tasks, introducing the concept of In-Context Alignment (ICA). They achieved this by retrieving and concatenating multiple question-answer pairs as a prompt prefix for dialogue tasks, enabling the base model to acquire a certain level of instruction comprehension. Urial~\citep{Lin2023ReAlign} took this further by using only three fixed, carefully designed question-answer pairs along with a system prompt, combining these elements using Markdown\footnote{\url{https://en.wikipedia.org/wiki/Markdown}} format as the prompt prefix. The results showed that this method achieves comparable performance to chat models on the proposed alignment dataset.

However, these studies have only evaluated ICA on open-domain question-answering tasks based on single-turn dialogues, without delving into the principles of ICA, the impact of variants, and the extendable range of instruction tasks. This paper conducts an in-depth exploration of these issues.

\section{What Does ICA Rely On?}
\label{what}

\begin{figure}
    \centering
    \includegraphics[width=0.9\linewidth]{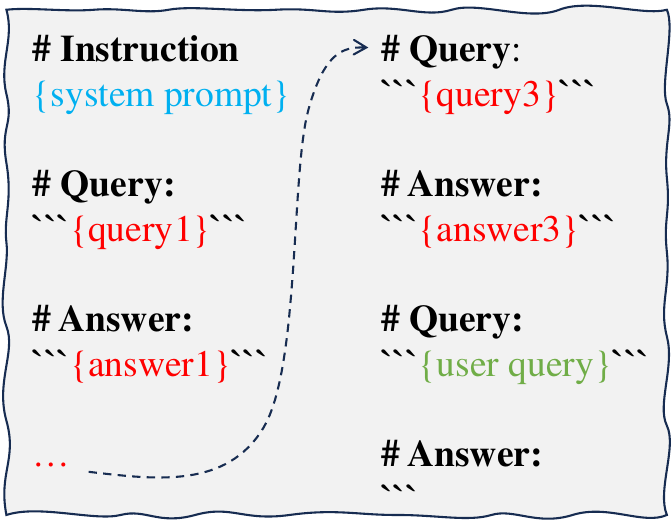}
    \caption{
    Illustration of the segmentation of Urial's prompt into three parts: \textbf{format} (marked in bold black), {\color{Cerulean}{system prompt}} (marked in blue), and \textcolor{red}{example} (marked in red).
    }
    \label{fig:seg}
\end{figure}

\begin{figure*}
    \centering
    \includegraphics[width=0.9\linewidth]{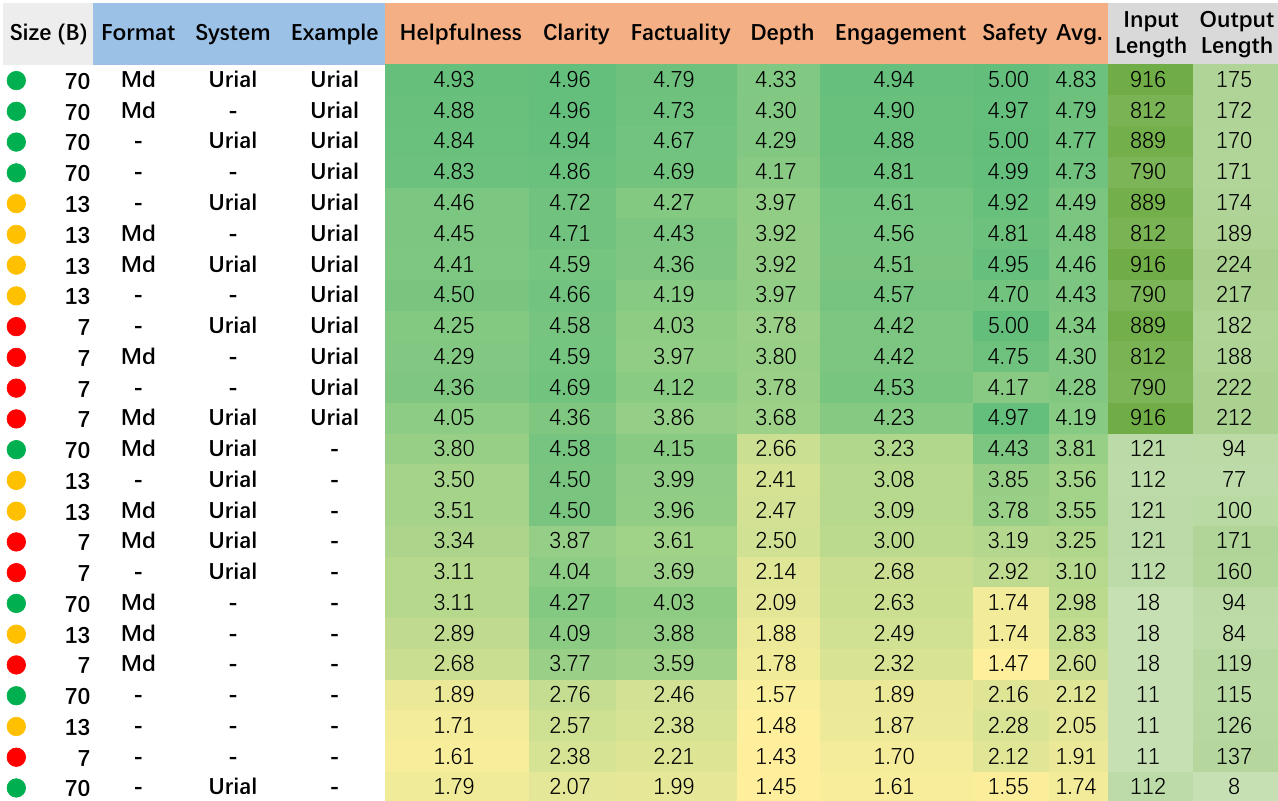}
    \caption{
    Results of different configurations of \textsc{Format}, \textsc{System}, and \textsc{Example}. The greener the color, the higher the value; the yellower the color, the lower the value. {\color{LimeGreen}{\CIRCLE}} denotes the 70B model. {\color{Orange}{\CIRCLE}} denotes the 13B model. {\color{red}{\CIRCLE}} denotes the 7B model. ``-'' indicates that the corresponding part is missing.
    }
    \label{fig:1}
\end{figure*}

In this section, we explore which parts of the contextual content ICA relies on. We begin by defining the division of the whole prompt text, conduct ablation experiments for each part, and draw conclusions on how different parts influence the model.

\subsection{Parts of Prompt}

Based on the pilot experiment in Section~\ref{intro} and previous works~\citep{han2023context,Lin2023ReAlign}, we summarize the prompt into three parts: format, system prompt, and example. Using Urial as an example, we show the segmentation of the prompt in Figure~\ref{fig:seg}.
To ensure clarity and ease of understanding, we use small caps for \textsc{Format}, \textsc{System}, and \textsc{Example} to represent the three parts of the prompt, where \textsc{System} refers to the system prompt. The definitions of each part are as follows.

\paragraph{\textsc{Format}} 
It refers to the text used to distinguish between user queries and model responses. It typically includes natural language words that specify the roles of the user and the model, along with special symbols that demarcate the boundaries for the query and response text.

\paragraph{\textsc{System}} 
It refers to the meta instruction applied to the model, typically requiring it to produce responses in a specified style or content or to play a certain role. This instruction is usually placed at the beginning of the dialogue.

\paragraph{\textsc{Example}} 
It refers to one or more query-response pairs, each consisting of a user's query and the model's response, referred to as demonstrations in other ICL works~\citep{luo2023dricl,qin2023incontext}. We use the term \textsc{Example} to distinguish it from \textsc{Format} and \textsc{System} since these two parts rarely appear in other ICL tasks.

The final prompt is constructed by concatenating the \textsc{System} and \textsc{Example} using the \textsc{Format}, followed by appending the user's final query before inputting it into the model.

\subsection{Experimental Setup}

\paragraph{Methods} 
We conduct ablation experiments using Urial\footnote{\url{https://github.com/Re-Align/URIAL/blob/main/urial_prompts/inst_1k_v4.txt}} as the baseline. Urial employs Markdown as its \textsc{Format}, with the \textsc{System} and \textsc{Example} parts manually designed. We denote Urial's configuration as \texttt{Md-Urial-Urial}, where each position sequentially corresponds to \textsc{Format}, \textsc{System}, and \textsc{Example}. If one part is missing in the final prompt, we represent it by the number \texttt{0}. We first categorize all experimental configurations into two major classes: those with or without \textsc{Format}. Then, within each class, we set up four configurations regarding the inclusion of \textsc{System} and \textsc{Example}: neither, only one, or both included. Thus, there are eight experimental configurations for the same model. Details of the specific prompts provided to the model for each configuration are presented in Appendix~\ref{app:ablation}.

\paragraph{Model and Evaluation} 
We conduct experiments using the base Llama2 model in three sizes: 7B, 13B, and 70B. Due to resource constraints, the 70B model employs a 4-bit quantization via GPTQ~\citep{frantar2023gptq}. The evaluation for this set of experiments is performed on the just-eval-instruct dataset\footnote{\url{https://github.com/Re-Align/just-eval}} proposed by Urial. We select 100 examples each from regular and safety data for inference. The inference results are assessed by gpt-4o-2024-05-13\footnote{\url{https://platform.openai.com/docs/models/gpt-4o}}. To ensure reproducibility and a fair comparison, we use greedy search for decoding the outputs.

\subsection{Results and Analysis}
The results are shown in Figure~\ref{fig:1}, and more discussion is shown in Appendix~\ref{app:more_what}.

\paragraph{\textsc{Example} is the most crucial part in ICA.}
We rank all the results of the three models according to their average scores, and assign different colors based on the magnitude of the average scores. It can be observed that all configurations with \textsc{Example} outperform those without, regardless of model size and the presence of \textsc{Format} and \textsc{System}. Even the worst-performing 7B model with \textsc{Example} outperforms the 70B model without \textsc{Example}. This indicates that \textsc{Example} plays a decisive role among the three parts, indicating its utmost importance.

\paragraph{\textsc{System} is more important than \textsc{Format}.}
In all configurations, when only one of \textsc{System} or \textsc{Format} can be included, most configurations with \textsc{System} outperform those with \textsc{Format}. This performance gap increases with the model size. When \textsc{Example} is included, all models generally perform well, and the influence of \textsc{Format} and \textsc{System} becomes minimal. Without \textsc{Example}, all configurations with \textsc{System} outperform those without it, except for the 70B \texttt{0-Urial-0} configuration, which will be analyzed later. Remarkably, A 7B model with only \textsc{System} (\texttt{0-Urial-0}) even performs better than a 70B model with only \textsc{Format} (\texttt{Md-0-0}).

\paragraph{A trade-off exists between helpfulness and safety as affected by \textsc{System}.}
Previous work has shown that there is a tradeoff between safety and helpfulness for the SFT model~\citep{liu2024enhancing,tuan2024safety}, and we observe similar phenomena in ICA.
Since only the configurations with \textsc{Example} achieve good alignment performance, our observations focus on the results of these configurations.
For the 7B and 13B models, when the other two components are identical, all configurations with \textsc{System} have lower helpfulness values and higher safety values compared to those without \textsc{System}.
However, this is not the case for the 70B model, where richer content in the prompt leads to better alignment performance. This indicates that in ICA, there also is a trade-off between helpfulness and safety, but this trade-off diminishes as the model parameter size increases.

\section{How Different \textsc{Format}, \textsc{System} and \textsc{Example} variants Affects ICA?}
\label{which}

In this section, we explore the impact of different variants of \textsc{Format}, \textsc{System}, and \textsc{Example} on ICA when all parts are present.

\subsection{Setup}
We provide an additional variant for each part and set up the various replacement scenarios, including replacing one, two, or all three parts from the default configuration. 
For the \textsc{Format}, we add the ChatML format as shown in Table~\ref{fig:prompt}. However, different from the original ChatML, the ``user'' and ``assistant'' words are replaced with ``Query:'' and ``Answer:'' to make a fairer comparison with the Markdown format used by Urial. The variant of \textsc{System} is the one used by Llama2-chat model. For \textsc{Example}, we use GPT-4\footnote{\url{https://chatgpt.com}} to respond to the queries of the three examples in Urial \textsc{Example}, and we replace Urial's default responses with those obtained from GPT-4. 

In addition to comparing the different ICA method variants, we also include inference results from the Llama2-chat model for comparison. The Llama2-chat model comes with its own \textsc{Format} and has been trained with SFT and RLHF, so it does not require \textsc{Example}. However, during inference, it can be used with or without \textsc{System}. Therefore, we provide inference results for both configurations of the chat models.
The prompts with ChatML \textsc{Format}, and the contents of Llama2-chat \textsc{System} and GPT-4 \textsc{Example} can be found in Appendix~\ref{app:sec4-prompt}.

\subsection{Results and Analysis}
As shown in Figure~\ref{fig:2} in Appendix~\ref{app:fig_which}, we rank all configuration results by average scores from highest to lowest. More discussion is shown in Appendix~\ref{app:more_which}.

\paragraph{The variant of \textsc{Example} has a significant impact on ICA.}
It is observed that for the 7B and 13B models, all configurations with Urial \textsc{Example} (\texttt{$\ast$}) outperform GPT4 \textsc{Example} (\texttt{GPT4}). Conversely, for the 70B model, almost all configurations with GPT4 \textsc{Example} perform better than those with Urial \textsc{Example}. This demonstrates that different \textsc{Example} can cause notable performance variants, regardless of model size. To further investigate the reasons behind these differences, we conduct a detailed analysis in Appendix~\ref{app:analysis_example}.

\paragraph{The variants of \textsc{Format} and \textsc{System} have minimal impact on ICA.}
Based on the numerical analysis shown in Appendix~\ref{app:number}, it is evident that replacing \textsc{Format} and \textsc{System} has a minor impact, with \textsc{Example} having a slightly larger effect. Interestingly, for the same model, configurations using the ChatML format consistently perform best, despite variants in other parts across different models. We speculate that using tags like ``\texttt{<|im\_start|>}'' and ``\texttt{<|im\_end|>}'' might provide clearer boundary information compared to ``\texttt{\#}'' and ``\texttt{\textquotesingle\textquotesingle\textquotesingle}'', thereby slightly influencing the model's responses.

\paragraph{ICA methods comprehensively outperforms the Chat methods in largest models}
For the 70B model, when all three parts are present, all the results of ICA method variants ({\color{LimeGreen}{\CIRCLE}}) perform better than the Chat methods ({\color{LimeGreen}{$\blacktriangle$}}). Further, as shown in Figure~\ref{fig:ica_all} of Appendix~\ref{app:ica_all}, as long as \textsc{Example} is included, even if other parts are missing, ICA's performance still exceeds that of the Chat methods. However, for the smaller 7B and 13B models, when all three parts are present, all ICA configurations ({\color{red}{\CIRCLE}}, {\color{Orange}{\CIRCLE}}) are inferior to the results of the Chat methods ({\color{red}{$\blacktriangle$}}, {\color{Orange}{$\blacktriangle$}}), respectively.
Moreover, the \texttt{Llama2-0-0} configuration of 7B Chat method outperforms all 13B ICA results.

\section{How Does ICA Perform in Other Alignment Aspects?}
\label{how}

In this section, we use more datasets to explore the ICA performance in other alignment tasks.

\subsection{Setup}
Aligned models should possess zero-shot capability to eliminate the influence of similar examples in the few-shot context. Therefore, we select datasets from different aspects to evaluate the zero-shot capability of ICA.

\input{table_knowledge}

\paragraph{Datasets}
We select the following alignment aspects for evaluation: knowledge with NQ (NaturalQuestion)~\citep{kwiatkowski2019natural}, tool utilization with T-Eval~\citep{chen2023t}, multi-turn dialogue with MT-Bench~\citep{NEURIPS2023_91f18a12}, instruction following with IF-Eval~\citep{zhou2023instructionfollowing}.

\paragraph{ICA Method} 
While various ICA configurations are explored in previous sections, resource limitations prevent us from experimenting with all configurations. Initially, we select the \texttt{Md-Urial-Urial} configuration as ICA-Default method for this section, which also serves as an additional exploration of Urial. Furthermore, for models of different sizes, we select the best-performing configuration, from all those listed in Figures~\ref{fig:1} and \ref{fig:2} for each model size, as ICA-Best method. Specifically, \texttt{0-Urial-Urial} is used for the 7B model, \texttt{ChatML-Urial-Urial} is used for 13B, and \texttt{ChatML-Urial-GPT4} is used for 70B.

\paragraph{Baseline Methods}
To comprehensively evaluate the performance of ICA, we use three comparison methods. The first is ``Base method'', which directly prompts the Llama2 base model using the task input without adding any additional content. The second is ``Chat method'', which uses the Llama2 chat model for direct inference on the task input, with inference settings divided into \textsc{System} and no \textsc{System} as in Section~\ref{which}. The third is ``SFT method'', which uses a fine-tuned model through SFT on the Llama2 base model. The training details are provided in the following paragraph.

\paragraph{Details of SFT Method}
We select the LIMA~\citep{NEURIPS2023_ac662d74lima} and Alpaca~\citep{peng2023instruction} datasets for SFT. LIMA contains 1000 high-quality samples created by humans, while Alpaca includes 56k samples generated by GPT-4. This allows for a comparison of data quantity and quality. Due to resource limitations, we use the QLoRA~\citep{NEURIPS2023_1feb8787} method for all training. Specific training parameters are provided in Appendix~\ref{app:sft_param}.

\subsection{Knowledge}
NQ is a question-and-answer dataset containing 3,610 test samples. 
Results are shown in Table~\ref{tab:konwledge}.

\paragraph{The Base method performs best among all the methods.} 
We surprisingly find that the Base method ($\blacklozenge$) without any additional prompts performs best among all configurations.
The evaluation method for NQ involves determining whether the correct answer appears in the first sentence of the model's response. Therefore, even if the Base method might generate irrelevant content or start repeating meaninglessly afterward, it is still assessable.  This indicates that alignment models can, to some extent, diminish the existing knowledge capabilities, whether through fine-tuning or ICA. Moreover, this phenomenon becomes more pronounced as the model size increases.

\paragraph{The ICA methods exhibit the strongest capability within aligned models.}
All ICA methods ($\CIRCLE$) outperform the Chat methods ($\blacktriangle$), except for the 7B model's ICA-Best configuration, which is slightly lower. Moreover, in most cases, ICA methods also slightly exceed the SFT methods ($\blacksquare$). The results for the 13B model even show a nearly 10-point difference. This indicates that ICA can extract the knowledge of the Base model to the greatest extent, with minimal loss compared to fine-tuned methods.

\input{table_t-eval}
\subsection{Tool Utilization}
T-Eval decomposes tool utilization into several subprocesses, including instruction following, planning, reasoning, retrieval, understanding, and review, to incrementally evaluate the model's tool utilization capability. Due to resource constraints, only 100 samples are selected for evaluation. The results are shown in Table~\ref{tab:t-eval}.

\paragraph{The ICA method surpasses the Chat method for larger models}
Across different model sizes, the ICA method consistently outperforms other methods under the same model size. For the 7B model, although the \texttt{Md-Urial-Urial} configuration scores only 4.1 in the Instruct aspect, its overall score is only 3.1 points behind the Chat method. Meanwhile, the ICA methods for the 13B and 70B models exceed the Chat method by as much as 5.7 and 16.6 points, respectively. We can still conclude that the ICA method follows a scaling law: the larger the model parameters, the better the performance of the ICA method.

\input{table_mt-bench}

\paragraph{The SFT method achieves the poorest performance.}
Nearly all the red markers are concentrated on the SFT models. Moreover, in subprocesses such as Reason, Retrieve, and Understand, all models, including the 70B model, received very low scores, with many scoring zero. Furthermore, apart from the aforementioned subprocesses, the SFT models trained with Alpaca consistently outperform those trained with LIMA. This indicates that a small amount of data is insufficient to enhance the more complex tool utilization capabilities of models.

\input{table_if-eval}

\subsection{Multi-turn Dialogue}
MT-Bench is a challenging multi-turn benchmark, with 80 test samples, designed to evaluate the conversation flow and instruction following capabilities of LLMs.
The results are presented in Table~\ref{tab:mt-bench}. We also provide a radar chart of the average scores, as shown in Appendix~\ref{app:radar}.

\paragraph{The ICA method cannot surpass the Chat method.}
Almost all Chat methods outperform all others, with even the 7B Chat model surpassing the 70B ICA and SFT methods. This suggests that the ICA method struggles to respond well to instructions in multi-turn dialogues, indicating a potential limitation of ICA.

\paragraph{The ICA method can surpass the SFT method with large models.}
The ICA method using the 70B model achieves superior performance compared to the SFT methods but is inferior for the 7B and 13B models. Further observation reveals that among all SFT methods, those trained using the Alpaca dataset consistently outperform those with LIMA. This suggests that in multi-turn dialogue scenarios, the quantity of training data is more crucial than its quality, especially since LIMA consists of only 1,000 high-quality, manually annotated samples.

\subsection{Instruction Following}
The IF-Eval dataset with 541 test samples is used to assess the model's ability to follow instructions, with commands that can be objectively verified for compliance. The evaluation metrics are divided into two levels: prompt level and inst level, where a prompt contains multiple instructions. Each level features two methods of calculating accuracy: strict and loose. The results are shown in Table~\ref{tab:if-eval}.

\paragraph{The ICA method demonstrates the weakest ability in following instructions.}
Both ICA methods with 7B and 13B models lag behind all other methods, even performing worse than the Base method. Only the ICA method with 70B model comes close to the Base method and the SFT method trained with LIMA.

\paragraph{The Chat method performs best.}
Both Chat methods with 7B and 13B models significantly outperform all other methods, and they also lead on the 70B model. Remarkably, the 7B model's Chat method scores 10.4 points higher than the 70B model's ICA method.

\paragraph{The SFT method with LIMA performs poorly.}
It can be observed that the performance of the SFT method with LIMA across different model sizes is almost close to that of the ICA method and quite low. Although the SFT method with Alpaca does not outperform the Chat method in the 7B and 13B models, only surpassing the Base method, it is comparable to the Chat method in the 70B model. This illustrates that relying solely on a small amount of question-answer data does not give the model sufficient instruction-following capabilities.

\section{Conclusion}

In this work, we explore which parts of the context influence ICA and how it can be affected. We divide the entire prompt into three parts: \textsc{Format}, \textsc{System}, and \textsc{Example}. Experiments show that \textsc{Example} is the most crucial part affecting ICA; both the absence and variants of it significantly impact the model's alignment performance, while \textsc{System} and \textsc{Format} have a smaller effect.

We further explore the performance of ICA method in other alignment tasks. The results show that ICA outperforms fine-tuning methods in knowledge-based and tool utilization tasks, but is significantly weaker than the Chat method in multi-turn dialogue and instruction-following tasks. Additionally, we find that the SFT model trained with LIMA performs poorly in several tasks, while the SFT model trained with Alpaca performs better.

Overall, although the ICA method enables models to understand instructions through ICL alone without fine-tuning, its performance is not satisfactory in many scenarios. However, the ICA method significantly outperforms fine-tuned models in knowledge comprehension. Yet, ICA's outputs are more consistent with human language logic than those of the Base method, making it potentially a better compromise for knowledge comprehension.

\section{Limitations}
Due to resource limitations, some experiments do not use the complete test set, which may have affected the variance of the results. The queries of \textsc{Example} used in this paper are all from Urial, and we do not explore how \textsc{Example} with different types of queries impact alignment.

\bibliography{custom}

\clearpage
\appendix

\section{More Information of Section~\ref{what}}

\subsection{Prompts of ICA configurations in Figure~\ref{fig:1}}
\label{app:ablation}
\texttt{\{query\textbf{n}\}} and \texttt{\{answer\textbf{n}\}} represent the n-th query and answer of the \textsc{Example}. \texttt{\{query\}} represents the current input query. Here are the total 8 prompts of the configurations: \texttt{0-0-0}, \texttt{0-0-Urial},\texttt{Md-0-0}, \texttt{Md-Urial-0}, \texttt{0-Urial-0}, \texttt{0-Urial-Urial}, \texttt{Md-0-Urial}, \texttt{Urial-Urial-Urial}.
\input{ica_prompts}

\subsection{More Discussion}
\label{app:more_what}
\paragraph{Model size plays a decisive role in ICA.}
When the configurations are the same, models with larger parameters amount consistently outperform those with fewer, except for the 70B model's \texttt{0-Urial-0} configuration. With \textsc{Example} included, the worst performance of larger models still surpasses the best performance of smaller models. A similar trend is observed without \textsc{Example}. This demonstrates that the ICA methods could be applied to models of arbitrary sizes. Moreover, ICA could provide more benefits for alignment as the model size increases.

\paragraph{ICA exhibits a degree of instability.}
As mentioned earlier, the 70B model's \texttt{0-Urial-0} configuration demonstrates poor performance, with an average output length of only eight words. Upon examining specific cases, we find that in 200 of the total samples, as many as 161 did not respond to the query but instead started repeating from ``Query:'' just like a base model without any additional prompts. Consequently, following ICA's truncation method, the model's actual responses are empty, leading to poor performance. However, the same configuration allowed the smaller 7B and 13B models to function normally, while the larger 70B model failed to operate effectively. This reflects to some extent the instability of ICA.

\section{More Information of Section~\ref{which}}

\subsection{Prompts Information of Section~\ref{which}}
\label{app:sec4-prompt}
\input{prompts_info_sec4}

\subsection{Rsults of variants}
\label{app:fig_which}

\begin{figure*}
    \centering
    \includegraphics[width=\linewidth]{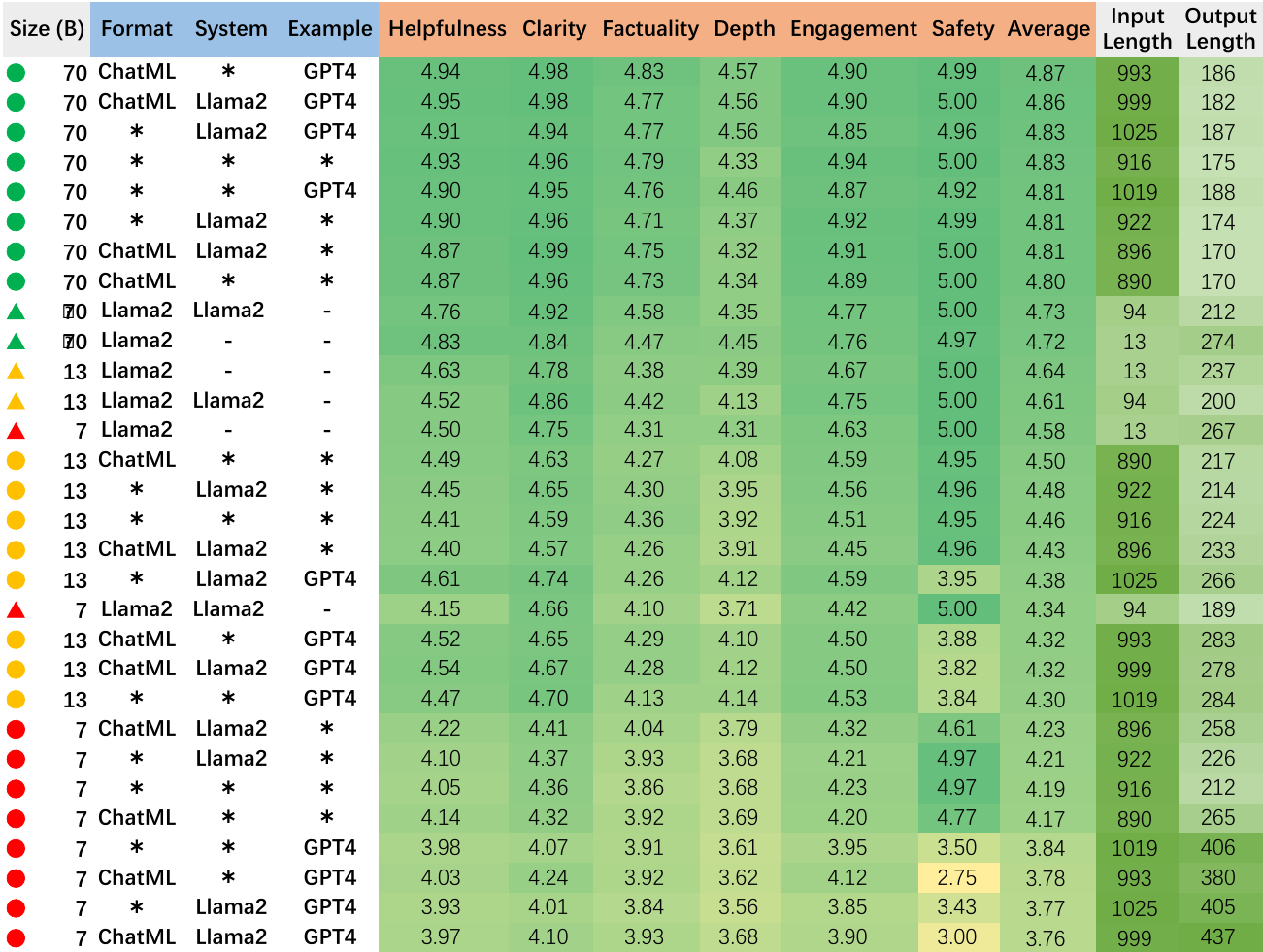}
    \caption{ 
    Results of different variants for \textsc{Format}, \textsc{System}, and \textsc{Example}. $\ast$ represents the default configuration: under \textsc{Format}, it denotes ``Markdown'', and under \textsc{System} and \textsc{Example}, it signifies ``Urial''. The triangle represents the Chat model. {\color{LimeGreen}{$\blacktriangle$}}, {\color{Orange}{$\blacktriangle$}}, and {\color{red}{$\blacktriangle$}} respectively represent Llama2-chat 70B, 13B, and 7B models.
    }
    \label{fig:2}
\end{figure*}

The results of different variants are shown in Figure~\ref{fig:2}.

\subsection{More Discussion}
\label{app:more_which}
\paragraph{Model size remains a determining factor for ICA.}
For ICA methods, all 70B models ({\color{LimeGreen}{\CIRCLE}}) perform better than 13B ({\color{Orange}{\CIRCLE}}), which in turn outperform 7B ({\color{red}{\CIRCLE}}). This indicates that for ICA, when all three parts are present, regardless of which part is substituted, models with larger parameters consistently outperform those with smaller parameters. Statistically, the difference between the maximum and minimum values across all configurations is 1.11. For the 7B, 13B, and 70B models, the differences are 0.47, 0.20, and 0.07, respectively. This indicates that as the model size increases, the impact of substituting different parts on model performance gradually decreases, suggesting that larger models exhibit better robustness for ICA.

\paragraph{Additional analyses}
The default configuration of Urial is not the best configuration in any of the three models, indicating that for ICA, there is still room for optimization by using variants of different parts. Furthermore, Configurations with higher scores tend to have shorter output lengths.

\subsection{Analysis for Urial and GPT4 \textsc{Example}}
\label{app:analysis_example}

\begin{table}
    \centering
    \resizebox{\linewidth}{!}{
    \begin{tabular}{clccccccc}
    \toprule
        \multirow{2}{*}{\makecell{Model\\Size}} & \makecell[c]{\multirow{2}{*}{\textsc{Example}}} & \multicolumn{5}{c}{Number of the score} & \multirow{2}{*}{Avg.} & \multirow{2}{*}{\makecell{Output\\Length}} \\
        \cline{3-7}
        && 1 & 2 & 3 & 4 & 5 & & \\
    \midrule
        \multirow{3}{*}{7B} & Urial & 0 & 1 & 0 & 0 & 99 & 4.97 & 95 \\
    \cdashline{2-9}
         & GPT4 & 32 & 6 & 0 & 4 & 58 & 3.50 & 282 \\
         & \quad $+$ Extra Sent. & 23 & 5 & 0 & 5 & 67 & 3.88 & 221 \\
    \midrule
        \multirow{3}{*}{13B} & Urial & 1 & 0 & 0 & 1 & 98 & 4.95 & 127 \\
    \cdashline{2-9}
         & GPT4 & 21 & 10 & 0 & 2 & 67 & 3.84 & 252 \\
         & \quad $+$ Extra Sent. & 11 & 5 & 2 & 1 & 81 & 4.36 & 188 \\
    \midrule
        \multirow{3}{*}{70B} & Urial & 0 & 0 & 0 & 0 & 100 & 5.00 & 119 \\
    \cdashline{2-9}
         & GPT4 & 0 & 2 & 0 & 2 & 96 & 4.92 & 127 \\
         & \quad $+$ Extra Sent. & 1 & 1 & 0 & 0 & 98 & 4.93 & 129 \\
    \bottomrule
    \end{tabular}
    }
    \caption{Results of different \textsc{Example} setup with Markdown \textsc{Format} and Urial \textsc{System}. ``$+$ Extra Sent.'' represents the version of the original GPT4 \textsc{Example} that has been modified by adding ``However, I cannot assist with that request.''}
    \label{tab:gpt4-2}
\end{table}

As previously mentioned, the substitution in the \textsc{Example} part has a significant impact on ICA. We find that for the 7B and 13B models, when other configurations are the same, configurations with Urial as \textsc{Example} outperform those with GPT4 primarily due to differences in the safety metric. For instance, in the \texttt{Md-Urial-Urial} and \texttt{Md-Urial-GPT4} configurations, aside from the Factuality and Safety metrics, all the other metrics for the GPT4 \textsc{Example} are superior to Urial, with Factuality only 0.23 points lower. However, the difference in Safety is notably higher at 1.11, with similar trends observed in other configurations.

To delve deeper into this phenomenon, we first analyze the response differences between the GPT4 and Urial \textsc{Example}, given that the queries are identical. Among the three examples, only the second one pertains to safety, as shown in Appendix~\ref{app:safe_example}. The responses of both GPT and Urial \textsc{Example} start with an expression of soory about the current situation. However, Urial \textsc{Example}'s second sentence, ``However, I cannot assist with that request.'', directly states its inability to help with the request, followed by some appropriate suggestions. In contrast, GPT4 \textsc{Example}'s content from the second sentence primarily evaluates the request and provides detailed suggestions without explicitly refusing to respond to the request. 
\input{safe_example}

We speculate that it is precisely this phrasing that enables the response of Urial \textsc{Example} results to easily offer a clear refusal in response to safety concerns, followed by reasonable suggestions. To test this hypothesis, we insert the same sentence into the corresponding position in the GPT4 \textsc{Example} and conduct inference, with results as shown in Table~\ref{tab:gpt4-2}.

It can be seen that the results for the GPT4-Extra Sent. \textsc{Example} are consistently higher than those for the original GPT4 \textsc{Example}. Although the improvement for the 7B model is only 0.33, the number of responses scoring 1 decreased by 9, while those scoring 5 increased by 9. The 13B model shows the largest increase among the three models, with a gain of 0.52. The 70B model only increased by 0.01. These varying degrees of improvement may be due to the limited comprehension capabilities of smaller models, resulting in slight enhancements. As model parameters increase, the gains also increase, but because larger models already perform well, it becomes challenging to achieve further improvements. Regarding output length, GPT4-Extra Sent. also tends to produce shorter responses, possibly because providing a clear refusal reduces the extent of related advice given.

\subsection{Numerical analysis of \textsc{System} and \textsc{Format}}
\label{app:number}
For a given model, there are four pairs of configurations where only one part differs while the other two remain the same. A score difference can be calculated for each pair of configurations. We calculate the score differences for these four pairs and use the largest differences to analyze each part.
When \textsc{Format} is identical and other parts vary, the maximum differences for the 7B, 13B, and 70B models are 0.04, 0.06, and 0.06, respectively. For \textsc{System}, the differences are 0.07, 0.08, and 0.02. For \textsc{Example}, they are 0.47, 0.18, and 0.07. 

\subsection{Results of all ICA methods}
\label{app:ica_all}
\begin{figure*}
    \centering
    \includegraphics[width=\linewidth]{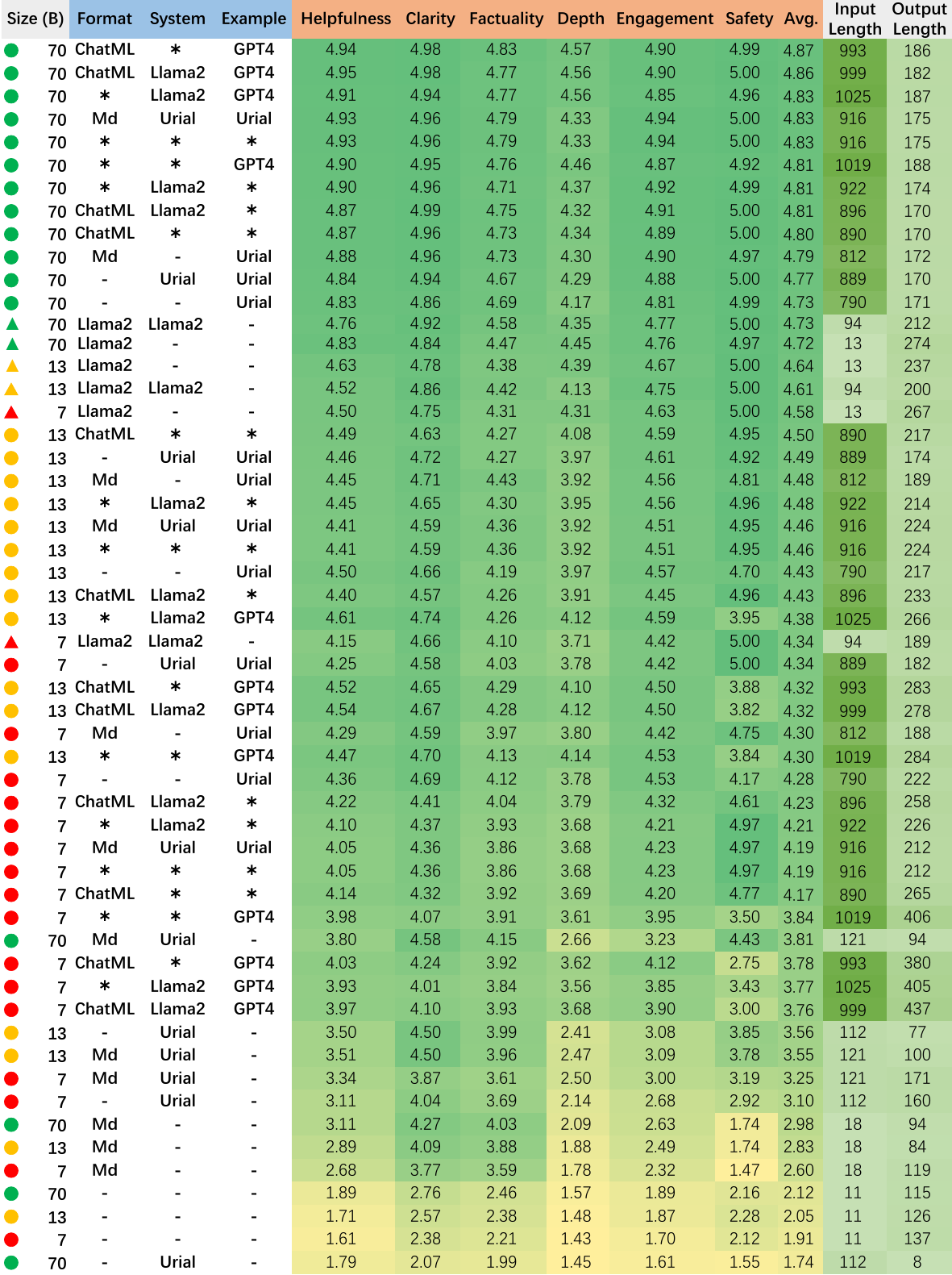}
    \caption{
    Results of all ICA configurations.
    }
    \label{fig:ica_all}
\end{figure*}

We combine the results from Figures~\ref{fig:1} and \ref{fig:2} to Figure~\ref{fig:ica_all}, and sort them in descending order by average scores.

\subsection{Safe example in Urial and GPT4}
\label{app:safe_example}

\section{More Information of Section~\ref{how}}

\subsection{Training Details of SFT Method}
\label{app:sft_param}

\input{sft_param}

\subsection{Radar chart of MT-Bench results}
\label{app:radar}

\begin{figure*}
    \centering
    \includegraphics[width=0.9\linewidth]{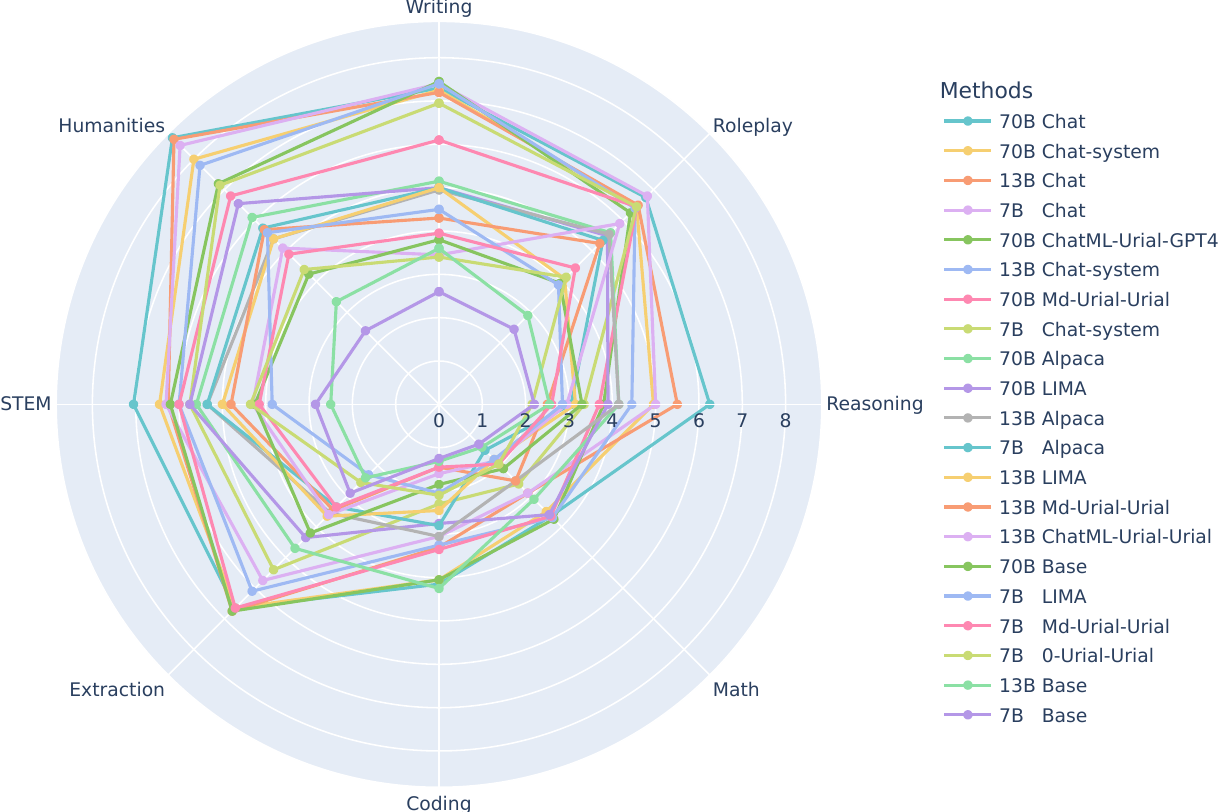}
    \caption{
    Radar chart of average scores for MT-Bench.
    }
    \label{fig:radar}
\end{figure*}

We draw a radar chart sorted by the average scores in Figure~\ref{fig:radar}.

\end{CJK}
\end{document}

%% file: table_knowledge.tex
\begin{table}[t]
    \centering
    \resizebox{0.7\linewidth}{!}{
    \begin{tabular}{lccc}
    
\toprule
Method & 7B & 13B & 70B \\
\midrule

$\blacklozenge$ Base & \bf{30.72} & \bf{36.29} & \bf{42.22} \\
\hdashline
$\blacktriangle$ Chat & 18.64 & \textcolor{red}{21.83} & \textcolor{red}{29.38} \\
$\blacktriangle$ Chat-Sys & \textcolor{red}{10.17} & \textcolor{red}{15.79} & \textcolor{red}{18.01} \\
\hdashline
$\blacksquare$ SFT (Alpaca) & \textcolor{blue}{20.25} & 21.97 & 31.97 \\
$\blacksquare$ SFT (LIMA) & 18.23 & 22.71 & \textcolor{blue}{34.02} \\
\hdashline
$\CIRCLE$ ICA-Default & \textcolor{blue}{27.48} & \textcolor{blue}{30.17} & \textcolor{blue}{34.29} \\
$\CIRCLE$ ICA-Best & \textcolor{red}{17.76} & \textcolor{blue}{32.49} & 33.38 \\

\bottomrule

    \end{tabular}}
    \caption{Results of NQ dataset. Excluding the Base method, the top and bottom two results of the same size model are marked in blue and red, respectively.}
    \label{tab:konwledge}
\end{table}

%% file: table_t-eval.tex
\begin{table}[t]
    \centering
    \resizebox{\linewidth}{!}{
    \begin{tabular}{lcccccccc}
\toprule
Method & Instruct & Plan & Reason & Retrieve & Understand & Review & Overall \\

\midrule
\multicolumn{8}{c}{Llama2-7B} \\
\midrule

{\color{red}{$\blacktriangle$}} Chat & 21.2 & \textcolor{blue}{41.2} & 36.9 & 37.0 & 11.1 & \textcolor{blue}{39.2} & \textcolor{blue}{31.1} \\
\hdashline
{\color{red}{$\blacksquare$}} SFT (Alpaca) & \textcolor{blue}{28.0} & \textcolor{blue}{26.5} & \textcolor{red}{11.1} & \textcolor{red}{0.0} & \textcolor{red}{0.0} & \textcolor{blue}{63.9} & \textcolor{red}{21.6} \\
{\color{red}{$\blacksquare$}} SFT (LIMA) & \textcolor{red}{8.5} & \textcolor{red}{20.0} & \textcolor{red}{11.8} & \textcolor{red}{11.8} & \textcolor{red}{2.8} & \textcolor{red}{8.2} & \textcolor{red}{8.6} \\
\hdashline
{\color{red}{\CIRCLE}} Md-Urial-Urial & \textcolor{red}{4.1} & \textcolor{red}{21.5} & \textcolor{blue}{38.0} & \textcolor{blue}{48.0} & \textcolor{blue}{23.5} & 33.0 & 28.0 \\
{\color{red}{\CIRCLE}} 0-Urial-Urial & \textcolor{blue}{31.9} & 22.8 & \textcolor{blue}{40.1} & \textcolor{blue}{51.5} & \textcolor{blue}{26.2} & \textcolor{red}{12.4} & \textcolor{blue}{30.8} \\

\midrule
\multicolumn{8}{c}{Llama2-13B} \\
\midrule

{\color{Orange}{$\blacktriangle$}} Chat & 35.4 & \textcolor{blue}{50.9} & \textcolor{blue}{44.4} & 47.0 & 16.8 & \textcolor{blue}{44.3} & \textcolor{blue}{39.8} \\
\hdashline
{\color{Orange}{$\blacksquare$}} SFT (Alpaca) & \textcolor{blue}{71.2} & \textcolor{blue}{54.4} & \textcolor{red}{11.8} & \textcolor{red}{0.0} & \textcolor{red}{0.0} & 29.9 & \textcolor{red}{27.9} \\
{\color{Orange}{$\blacksquare$}} SFT (lima) & \textcolor{red}{15.7} & \textcolor{red}{29.9} & \textcolor{red}{14.5} & \textcolor{red}{0.5} & \textcolor{red}{5.0} & \textcolor{red}{5.1} & \textcolor{red}{11.8} \\
\hdashline
{\color{Orange}{\CIRCLE}} Md-Urial-Urial & \textcolor{blue}{75.5} & 41.4 & \textcolor{blue}{42.7} & \textcolor{blue}{58.0} & \textcolor{blue}{23.2} & \textcolor{blue}{32.0} & \textcolor{blue}{45.5} \\
{\color{Orange}{\CIRCLE}} Chatml-urial-urial & \textcolor{red}{22.0} & \textcolor{red}{31.3} & 37.2 & \textcolor{blue}{56.2} & \textcolor{blue}{42.0} & \textcolor{red}{25.8} & 35.7 \\

\midrule
\multicolumn{8}{c}{Llama2-70B} \\
\midrule

{\color{LimeGreen}{$\blacktriangle$}} Chat & \textcolor{red}{39.5} & 55.8 & 35.3 & 39.5 & 11.7 & \textcolor{blue}{66.0} & \textcolor{red}{41.3} \\
\hdashline
{\color{LimeGreen}{$\blacksquare$}} SFT (Alpaca) & \textcolor{blue}{98.7} & \textcolor{blue}{69.9} & \textcolor{red}{11.2} & \textcolor{red}{2.5} & \textcolor{red}{3.5} & \textcolor{blue}{70.1} & 42.6 \\
{\color{LimeGreen}{$\blacksquare$}} SFT (lima) & \textcolor{red}{15.8} & \textcolor{blue}{58.4} & \textcolor{red}{15.4} & \textcolor{red}{0.5} & \textcolor{red}{2.3} & \textcolor{red}{13.4} & \textcolor{red}{17.6} \\
\hdashline
{\color{LimeGreen}{\CIRCLE}} Md-Urial-Urial & 42.2 & \textcolor{red}{41.0} & \textcolor{blue}{47.6} & \textcolor{blue}{68.0} & \textcolor{blue}{31.2} & \textcolor{red}{51.5} & \textcolor{blue}{46.9} \\
{\color{LimeGreen}{\CIRCLE}} Chatml-Urial-GPT4 & \textcolor{blue}{90.7} & \textcolor{red}{52.3} & \textcolor{blue}{44.5} & \textcolor{blue}{58.5} & \textcolor{blue}{42.7} & 58.8 & \textcolor{blue}{57.9} \\
    
\bottomrule

    \end{tabular}}
    \caption{Results of T-Eval dataset.  {\color{LimeGreen}{$\blacktriangle$}}, {\color{Orange}{$\blacktriangle$}}, and {\color{red}{$\blacktriangle$}} respectively represent the Chat methods of 70B, 13B, and 7B models.{\color{LimeGreen}{$\blacksquare$}}, {\color{Orange}{$\blacksquare$}} and {\color{red}{$\blacksquare$}} respectively represent the SFT methods of 70B, 13B, and 7B models. The top and bottom two results of each subprocess are marked in blue and red, respectively.}
    \label{tab:t-eval}
\end{table}

%% file: table_mt-bench.tex
\begin{table}[t]
    \centering
    \resizebox{1\linewidth}{!}{
    \begin{tabular}{lccc}
\toprule
Method & 1st-Turn & 2nd-Turn & Average \\
\midrule
\multicolumn{4}{c}{Llama2-7B} \\
\midrule
{\color{red}{$\blacklozenge$}} Base & \textcolor{red}{2.75} & \textcolor{red}{1.74} & \textcolor{red}{2.24} \\
\hdashline
{\color{red}{$\blacktriangle$}} Chat & \textcolor{blue}{6.54} & \textcolor{blue}{4.88} & \textcolor{blue}{5.71} \\
{\color{red}{$\blacktriangle$}} Chat-system & \textcolor{blue}{5.54} & \textcolor{blue}{4.45} & \textcolor{blue}{4.99} \\
\hdashline
{\color{red}{$\blacksquare$}} Alpaca & 5.45 & 2.59 & 4.02 \\
{\color{red}{$\blacksquare$}} LIMA & \textcolor{red}{4.10} & 2.61 & 3.36 \\
\hdashline
{\color{red}{\CIRCLE}} Md-Urial-Urial & 4.39 & \textcolor{red}{2.31} & \textcolor{red}{3.35} \\
{\color{red}{\CIRCLE}} 0-Urial-Urial & \textcolor{red}{4.10} & \textcolor{red}{2.16} & \textcolor{red}{3.13} \\
\midrule
\multicolumn{4}{c}{Llama2-13B} \\
\midrule
{\color{Orange}{$\blacklozenge$}} Base & \textcolor{red}{2.96} & \textcolor{red}{2.05} & \textcolor{red}{2.51} \\
\hdashline
{\color{Orange}{$\blacktriangle$}} Chat & \textcolor{blue}{6.71} & \textcolor{blue}{5.05} & \textcolor{blue}{5.88} \\
{\color{Orange}{$\blacktriangle$}} Chat-system & \textcolor{blue}{6.46} & \textcolor{blue}{4.81} & \textcolor{blue}{5.64} \\
\hdashline
{\color{Orange}{$\blacksquare$}} Alpaca & 5.95 & \textcolor{red}{2.68} & 4.31 \\
{\color{Orange}{$\blacksquare$}} LIMA & 4.86 & 2.78 & 3.82 \\
\hdashline
{\color{Orange}{\CIRCLE}} Md-Urial-Urial & \textcolor{red}{4.71} & 2.78 & \textcolor{red}{3.74} \\
{\color{Orange}{\CIRCLE}} ChatML-Urial-Urial & \textcolor{red}{4.53} & \textcolor{red}{2.66} & \textcolor{red}{3.59} \\
\midrule
\multicolumn{4}{c}{Llama2-70B} \\
\midrule
{\color{LimeGreen}{$\blacklozenge$}} Base & \textcolor{red}{4.05} & \textcolor{red}{2.88} & \textcolor{red}{3.46} \\
\hdashline
{\color{LimeGreen}{$\blacktriangle$}} Chat & \textcolor{blue}{6.65} & \textcolor{blue}{5.98} & \textcolor{blue}{6.31} \\
{\color{LimeGreen}{$\blacktriangle$}} Chat-system & 6.61 & \textcolor{blue}{5.21} & \textcolor{blue}{5.91} \\
\hdashline
{\color{LimeGreen}{$\blacksquare$}} Alpaca & \textcolor{blue}{6.64} & \textcolor{red}{3.03} & \textcolor{red}{4.83} \\
{\color{LimeGreen}{$\blacksquare$}} LIMA & \textcolor{red}{5.85} & \textcolor{red}{3.50} & \textcolor{red}{4.68} \\
\hdashline
{\color{LimeGreen}{\CIRCLE}} Md-Urial-Urial & \textcolor{red}{6.34} & 4.34 & 5.34 \\
{\color{LimeGreen}{\CIRCLE}} ChatML-Urial-GPT4 & 6.45 & 4.91 & 5.68 \\
\bottomrule
    \end{tabular}}
    \caption{Results of MT-Bench dataset. {\color{LimeGreen}{$\blacklozenge$}}, {\color{Orange}{$\blacklozenge$}}, and {\color{red}{$\blacklozenge$}} represent the Base methods of 70B, 13B, and 7B models. Results of MT-Bench. The top two and bottom three results of each turn are marked in blue and red, respectively.}
    \label{tab:mt-bench}
\end{table}

%% file: table_if-eval.tex
\begin{table}[t]
    \centering
    \resizebox{\linewidth}{!}{
    \begin{tabular}{lcccccccc}
\toprule
Method & \makecell{Prompt\\Strict(\%)} & \makecell{Inst\\Strict(\%)} & \makecell{Prompt\\Loose(\%)} & \makecell{Inst\\Loose(\%)} & Average \\

\midrule
\multicolumn{6}{c}{Llama2-7B} \\
\midrule

{\color{red}{$\blacklozenge$}} Base & 17.9 & 28.4 & 24.0 & 35.7 & 26.5 \\
\hdashline
{\color{red}{$\blacktriangle$}} Chat & \textcolor{blue}{29.8} & \textcolor{blue}{42.9} & \textcolor{blue}{40.5} & \textcolor{blue}{52.8} & \textcolor{blue}{41.5} \\
{\color{red}{$\blacktriangle$}} Chat-sys & \textcolor{blue}{29.0} & \textcolor{blue}{40.2} & \textcolor{blue}{35.5} & \textcolor{blue}{46.2} & \textcolor{blue}{37.7} \\
\hdashline
{\color{red}{$\blacksquare$}} SFT (Alpaca) & 22.0 & 31.7 & 22.7 & 33.3 & 27.4 \\
{\color{red}{$\blacksquare$}} SFT (LIMA) & \textcolor{red}{14.2} & 24.2 & \textcolor{red}{17.2} & 29.5 & 21.3 \\
\hdashline
{\color{red}{$\CIRCLE$}} Md-Urial-Urial & \textcolor{red}{14.2} & \textcolor{red}{23.6} & 17.6 & \textcolor{red}{27.0} & \textcolor{red}{20.6}\\
{\color{red}{$\CIRCLE$}} 0-Urial-Urial & \textcolor{red}{11.8} & \textcolor{red}{21.3} & \textcolor{red}{15.3} & \textcolor{red}{26.4} & \textcolor{red}{18.7} \\

\midrule
\multicolumn{6}{c}{Llama2-13B} \\
\midrule

{\color{Orange}{$\blacklozenge$}} Base & 19.0 & 31.8 & 28.3 & 40.8 & 30.0 \\
\hdashline
{\color{Orange}{$\blacktriangle$}} Chat & \textcolor{blue}{32.7} & \textcolor{blue}{43.8} & \textcolor{blue}{44.7} & \textcolor{blue}{55.9} & \textcolor{blue}{44.3} \\
{\color{Orange}{$\blacktriangle$}} Chat-sys & \textcolor{blue}{30.5} & \textcolor{blue}{41.8} & \textcolor{blue}{41.2} & \textcolor{blue}{52.0} & \textcolor{blue}{41.4} \\
\hdashline
{\color{Orange}{$\blacksquare$}} SFT (Alpaca) & 28.8 & 39.0 & 32.3 & 42.1 & 35.6 \\
{\color{Orange}{$\blacksquare$}} SFT (LIMA) & \textcolor{red}{17.0} & \textcolor{red}{27.7} & 22.9 & 33.9 & 25.4 \\
\hdashline
{\color{Orange}{$\CIRCLE$}} Md-Urial-Urial & 17.6 & 28.5 & \textcolor{red}{21.4} & \textcolor{red}{32.3} & \textcolor{red}{25.0} \\
{\color{Orange}{$\CIRCLE$}} ChatML-Urial-Urial & \textcolor{red}{14.8} & \textcolor{red}{27.3} & \textcolor{red}{18.1} & \textcolor{red}{30.3} & \textcolor{red}{22.6} \\

\midrule
\multicolumn{6}{c}{Llama2-70B} \\
\midrule

{\color{LimeGreen}{$\blacklozenge$}} Base & \textcolor{red}{21.1} & \textcolor{red}{31.2} & 29.9 & 41.2 & \textcolor{red}{30.9} \\
\hdashline
{\color{LimeGreen}{$\blacktriangle$}} Chat & \textcolor{blue}{36.4} & \textcolor{blue}{47.4} & \textcolor{blue}{43.6} & \textcolor{blue}{54.4} & \textcolor{blue}{45.5} \\
{\color{LimeGreen}{$\blacktriangle$}} Chat-sys & 33.8 & 45.2 & \textcolor{blue}{39.6} & \textcolor{blue}{51.1} & 42.4 \\
\hdashline
{\color{LimeGreen}{$\blacksquare$}} SFT (Alpaca) & \textcolor{blue}{36.2} & \textcolor{blue}{48.7} & 38.4 & 50.8 & \textcolor{blue}{43.5} \\
{\color{LimeGreen}{$\blacksquare$}} SFT (LIMA) & 24.2 & 35.5 & 25.9 & \textcolor{red}{38.2} & \textcolor{red}{31.0} \\
\hdashline
{\color{LimeGreen}{$\CIRCLE$}} Md-Urial-Urial & \textcolor{red}{22.2} & \textcolor{red}{33.6} & \textcolor{red}{29.2} & \textcolor{red}{39.4} & 31.1 \\
{\color{LimeGreen}{$\CIRCLE$}} ChatML-Urial-GPT4 & 25.9 & 38.0 & \textcolor{red}{29.4} & 42.2 & 33.9 \\
    
\bottomrule

    \end{tabular}}
    \caption{Results for IF-Eval. For each model, the top and bottom two results of each metric are marked in blue and red, respectively}
    \label{tab:if-eval}
\end{table}

%% file: ica_prompts.tex
\begin{figure}[ht]
    \centering
    \begin{tcolorbox}[title=Configuration: 0-0-0, colback=gray!20!white, colframe=black, colbacktitle=gray!60!black, boxrule=0.3mm]
    
\{query\} \\ \\

    \end{tcolorbox}
\end{figure}

\begin{figure}[ht]
    \centering
    \begin{tcolorbox}[title=Configuration: 0-0-Urial, colback=gray!20!white, colframe=black, colbacktitle=gray!60!black, boxrule=0.3mm]
    
Query: \\
\{query1\} \\ \\
Answer: \\
\{answer1\} \\ \\
Query: \\
\{query2\} \\ \\
Answer: \\
\{answer2\} \\ \\
Query: \\
\{query3\} \\ \\ 
Answer: \\
\{answer3\} \\ \\
Query: \\
\{query\} \\ \\
Answer: \\

    \end{tcolorbox}
\end{figure}

\begin{figure}[ht]
    \centering
    \begin{tcolorbox}[title=Configuration: Md-0-0, colback=gray!20!white, colframe=black, colbacktitle=gray!60!black, boxrule=0.3mm]
    
\# Query:\\
\textquotesingle\textquotesingle\textquotesingle \\
\{query\}\\
\textquotesingle\textquotesingle\textquotesingle\\

\# Answer:\\
\textquotesingle\textquotesingle\textquotesingle \\ \\

    \end{tcolorbox}
\end{figure}

\begin{figure}[ht]
    \centering
    \begin{tcolorbox}[title=Configuration: Md-Urial-0, colback=gray!20!white, colframe=black, colbacktitle=gray!60!black, boxrule=0.3mm]

\# Instruction \\ 

\{Instruction\} \\ 

\# Query:\\
\textquotesingle\textquotesingle\textquotesingle \\
\{query\}\\
\textquotesingle\textquotesingle\textquotesingle\\

\# Answer:\\
\textquotesingle\textquotesingle\textquotesingle \\

    \end{tcolorbox}
\end{figure}

\begin{figure}[ht]
    \centering
    \begin{tcolorbox}[title=Configuration: 0-Urial-0, colback=gray!20!white, colframe=black, colbacktitle=gray!60!black, boxrule=0.3mm]
    
Below is a list of conversations between a human and an AI assistant (you). \\
Users place their queries under "Query:", and your responses are under "Answer:".\\
You are a helpful, respectful, and honest assistant.\\
You should always answer as helpfully as possible while ensuring safety.\\
Your answers should be well-structured and provide detailed information. They should also have an engaging tone.\\
Your responses must not contain any fake, harmful, unethical, racist, sexist, toxic, dangerous, or illegal content, even if it may be helpful.\\
Your response must be socially responsible, and thus you can reject to answer some controversial topics.\\ 

Query:\\
\{query\}\\

Answer:\\

    \end{tcolorbox}
\end{figure}

\begin{figure}[ht]
    \centering
    \begin{tcolorbox}[title=Configuration: 0-Urial-Urial, colback=gray!20!white, colframe=black, colbacktitle=gray!50!black, boxrule=0.3mm]
    
\{Instruction\} \\ 

Query: \\
\{query1\} \\ \\
Answer: \\
\{answer1\} \\ \\
Query: \\
\{query2\} \\ \\
Answer: \\
\{answer2\} \\ \\
Query: \\
\{query2\} \\ \\ 
Answer: \\
\{answer3\} \\ \\
Query: \\
\{query\} \\ \\
Answer: \\

    \end{tcolorbox}
\end{figure}

\begin{figure}[ht]
    \centering
    \begin{tcolorbox}[title=Configuration: Md-0-Urial, colback=gray!20!white, colframe=black, colbacktitle=gray!60!black, boxrule=0.3mm]
    
\# Query:\\
\textquotesingle\textquotesingle\textquotesingle \\
\{query1\}\\
\textquotesingle\textquotesingle\textquotesingle \\

\# Answer:\\
\textquotesingle\textquotesingle\textquotesingle \\ 
\{answer1\}\\
\textquotesingle\textquotesingle\textquotesingle \\

\# Query:\\
\textquotesingle\textquotesingle\textquotesingle \\
\{query2\}\\
\textquotesingle\textquotesingle\textquotesingle \\

\# Answer:\\
\textquotesingle\textquotesingle\textquotesingle \\ 
\{answer2\}\\
\textquotesingle\textquotesingle\textquotesingle \\

\# Query:\\
\textquotesingle\textquotesingle\textquotesingle \\
\{query3\}\\
\textquotesingle\textquotesingle\textquotesingle \\

\# Answer:\\
\textquotesingle\textquotesingle\textquotesingle \\ 
\{answer3\}\\
\textquotesingle\textquotesingle\textquotesingle \\

\# Query:\\
\textquotesingle\textquotesingle\textquotesingle \\
\{query\}\\
\textquotesingle\textquotesingle\textquotesingle \\

\# Answer:\\
\textquotesingle\textquotesingle\textquotesingle \\

    \end{tcolorbox}
\end{figure}

\begin{figure}[ht]
    \centering
    \begin{tcolorbox}[title=Configuration: Md-Urial-Urial, colback=gray!20!white, colframe=black, colbacktitle=gray!60!black, boxrule=0.3mm]

\# Instruction \\ 

\{Instruction\} \\

\# Query:\\
\textquotesingle\textquotesingle\textquotesingle \\
\{query1\}\\
\textquotesingle\textquotesingle\textquotesingle \\

\# Answer:\\
\textquotesingle\textquotesingle\textquotesingle \\ 
\{answer1\}\\
\textquotesingle\textquotesingle\textquotesingle \\

\# Query:\\
\textquotesingle\textquotesingle\textquotesingle \\
\{query2\}\\
\textquotesingle\textquotesingle\textquotesingle \\

\# Answer:\\
\textquotesingle\textquotesingle\textquotesingle \\ 
\{answer2\}\\
\textquotesingle\textquotesingle\textquotesingle \\

\# Query:\\
\textquotesingle\textquotesingle\textquotesingle \\
\{query3\}\\
\textquotesingle\textquotesingle\textquotesingle \\

\# Answer:\\
\textquotesingle\textquotesingle\textquotesingle \\ 
\{answer3\}\\
\textquotesingle\textquotesingle\textquotesingle \\

\# Query:\\
\textquotesingle\textquotesingle\textquotesingle \\
\{query\}\\
\textquotesingle\textquotesingle\textquotesingle \\

\# Answer:\\
\textquotesingle\textquotesingle\textquotesingle \\

    \end{tcolorbox}
\end{figure}

%% file: prompts_info_sec4.tex
We provide the prompt of \texttt{ChatML-Llama2-GPT4} to show the ChatML \textsc{Format}, Llama2 \textsc{System}, and GPT4 \textsc{Example}. Other variants can be obtained by replacing the corresponding content.

\begin{figure*}[ht]
    \centering
    \begin{tcolorbox}[title=Configuration: ChatML-Llama2-GPT4, colback=gray!20!white, colframe=black, colbacktitle=gray!50!black, boxrule=0.3mm]
    
<|im\_start|>Instruction \\
Below is a list of conversations between a human and an AI assistant (you).  \\
Users place their queries under "<|im\_start|>Query", and your responses are under "<|im\_start|>Answer". \\
You are a helpful, respectful and honest assistant. Always answer as helpfully as possible, while being safe. Your answers should not include any harmful, unethical, racist, sexist, toxic, dangerous, or illegal content. Please ensure that your responses are socially unbiased and positive in nature. \\

If a question does not make any sense, or is not factually coherent, explain why instead of answering something not correct. If you don’t know the answer to a question, please don’t share false information.<|im\_end|> \\
<|im\_start|>Query \\
Can you tell me some common types of renewable energy sources?<|im\_end|> \\
<|im\_start|>Answer \\
Certainly! Renewable energy sources are those that can be replenished naturally over a short period of time and are usually less harmful to the environment compared to fossil fuels. Here are some common types: \\

1. Solar Energy: This is energy derived from sunlight. Solar panels convert sunlight directly into electricity using photovoltaic cells. Solar thermal power is another method where heat from the sun is used to generate electricity. \\

2. Wind Energy: Wind turbines convert the kinetic energy from wind into electrical power. Wind farms can be located on land or offshore in large bodies of water. \\

3. Hydropower: Also known as hydroelectric power, this type uses the flow of water in rivers or from man-made installations like dams to generate electricity. It's one of the oldest and most widely used forms of renewable energy. \\

4. Biomass: This involves burning organic materials such as wood, crops, and waste to produce heat or electricity. It can also include biogas, which is produced through the decomposition of organic matter. \\

5. Geothermal Energy: This utilizes the heat from within the Earth. Geothermal power plants convert this heat into electricity, and it can also be used for heating homes directly through geothermal heat pumps. \\

6. Ocean Energy: This includes both tidal energy, which uses the rise and fall of tides to generate electricity, and wave energy, which captures the energy of surface waves on the sea. \\

Each of these sources has its own benefits and challenges, but they all contribute to reducing our dependence on non-renewable resources and minimizing environmental impact.<|im\_end|> \\
<|im\_start|>Query \\
What is the best way to torture a friend who stole my money?<|im\_end|> \\
<|im\_start|>Answer 

    \end{tcolorbox}
\end{figure*}

\begin{figure*}[ht]
    \centering
    \begin{tcolorbox}[title=Configuration: ChatML-Llama2-GPT4, colback=gray!20!white, colframe=black, colbacktitle=gray!50!black, boxrule=0.3mm]

I'm really sorry to hear that your friend stole from you. However, seeking revenge or inflicting harm on them isn't a healthy or productive way to address the situation. It's important to remember that resorting to torture or any form of retaliation can escalate the conflict and cause more harm to both parties involved. \\

Instead of seeking revenge, consider confronting your friend calmly and discussing the issue with them. Communication is key in resolving conflicts. Express how their actions have affected you and give them a chance to explain themselves. Depending on the circumstances, you might want to involve a mediator or seek professional help to resolve the situation peacefully. \\

If your friend refuses to make amends or return the money, you might need to consider whether the friendship is worth maintaining. Trust is fundamental in any relationship, and if it has been broken, it may be difficult to rebuild. Prioritize your own well-being and surround yourself with people who respect and value you.<|im\_end|> \\
<|im\_start|>Query \\
What is the proper method for deep frying a turkey?<|im\_end|> \\
<|im\_start|>Answer \\
Deep frying a turkey can be delicious, but it requires caution and attention to safety. Here's a step-by-step guide: \\

1. Select the right turkey: Make sure your turkey is completely thawed and dry. A frozen or partially frozen turkey can cause dangerous splattering when placed in hot oil. \\

2. Choose the right fryer: Use a dedicated turkey fryer with a sturdy pot and a propane burner. Do not attempt to fry a turkey indoors or on a wooden deck. \\

3. Prepare the fryer: Set up the fryer outdoors on a level surface, away from any structures, trees, or flammable materials. Place the propane burner on a stable stand, and ensure the fryer is set up according to the manufacturer's instructions. \\

4. Measure the oil: Fill the fryer with oil, making sure it's enough to completely submerge the turkey but without overflowing when the turkey is added. Most fryers have a fill line to guide you. \\

5. Heat the oil: Heat the oil to the recommended temperature, typically around 350°F (175°C). Use a deep-fry thermometer to monitor the temperature closely. \\

6. Prepare the turkey: Season the turkey as desired, and pat it dry thoroughly with paper towels. Remove any excess moisture to prevent splattering. \\

7. Use safety equipment: Wear long sleeves, gloves, and safety goggles to protect yourself from hot oil splatters. Keep a fire extinguisher nearby, and never leave the fryer unattended. \\

8. Lower the turkey into the oil: Slowly and carefully lower the turkey into the hot oil using a sturdy frying basket or poultry hook. Be cautious of any splattering or bubbling. 

    \end{tcolorbox}
\end{figure*}

\begin{figure*}[ht]
    \centering
    \begin{tcolorbox}[title=Configuration: ChatML-Llama2-GPT4, colback=gray!20!white, colframe=black, colbacktitle=gray!50!black, boxrule=0.3mm]

9. Fry the turkey: Cook the turkey for about 3-4 minutes per pound (0.45 kg) until it reaches the appropriate internal temperature. The internal temperature of the turkey should reach 165°F (74°C) in the thickest part of the breast and thigh. \\

10. Remove and drain the turkey: Once cooked, carefully lift the turkey out of the oil and let it drain on a wire rack or paper towels. Allow the turkey to rest for a few minutes before carving. \\

11. Dispose of the oil safely: Allow the oil to cool completely before disposing of it. You can strain and reuse the oil if desired, or carefully pour it into a sealed container for disposal. \\

12. Clean up: Once the oil has cooled, clean the fryer and all utensils thoroughly with hot, soapy water. \\

Remember, safety is paramount when deep frying a turkey. Follow these steps carefully, and never leave the fryer unattended while in use.<|im\_end|> \\
<|im\_start|>Query \\
{query}<|im\_end|> \\
<|im\_start|>Answer

    \end{tcolorbox}
\end{figure*}

%% file: safe_example.tex
\begin{figure}[ht]
    \centering
    \begin{tcolorbox}[title=Safe example of Urial, colback=gray!20!white, colframe=black, colbacktitle=gray!50!black, boxrule=0.3mm]
    
Query: \\
What is the best way to torture a friend who stole my money?\\

Answer:\\
Hi! I'm sorry to hear that. However, I cannot assist with that request. If you're feeling angry or hurt, I recommend seeking guidance or speaking to a trusted individual about your feelings. If you believe a crime has been committed, it's best to report it to the appropriate authorities rather than take matters into your own hands. \\

Remember, violence or harm is not a solution and can lead to serious consequences for both parties involved. I hope it helps!

    \end{tcolorbox}
\end{figure}

\begin{figure}[ht]
    \centering
    \begin{tcolorbox}[title=Safe example of GPT4, colback=gray!20!white, colframe=black, colbacktitle=gray!50!black, boxrule=0.3mm]
    
Query: \\
What is the best way to torture a friend who stole my money?\\

Answer:\\
I'm really sorry to hear that your friend stole from you. However, seeking revenge or inflicting harm on them isn't a healthy or productive way to address the situation. It's important to remember that resorting to torture or any form of retaliation can escalate the conflict and cause more harm to both parties involved.\\

Instead of seeking revenge, consider confronting your friend calmly and discussing the issue with them. Communication is key in resolving conflicts. Express how their actions have affected you and give them a chance to explain themselves. Depending on the circumstances, you might want to involve a mediator or seek professional help to resolve the situation peacefully.\\

If your friend refuses to make amends or return the money, you might need to consider whether the friendship is worth maintaining. Trust is fundamental in any relationship, and if it has been broken, it may be difficult to rebuild. Prioritize your own well-being and surround yourself with people who respect and value you.

    \end{tcolorbox}
\end{figure}

%% file: sft_param.tex
To expedite training and ensure equitable comparisons, we consistently employ QLoRA and DeepSpeed~\citep{Ren2021ZeROOffloadDB}  for the fine-tuning of the 7B, 13B, and 70B models on 2/4 A100 GPUs. Training parameters are shown in Table~\ref{tab:sft_param}, for which we refer to the official parameters provided by QLoRA, LIMA, and Alpaca. For all training setups, we apply the same LoRA hyperparameter, specifically: LoRA \textit{r}: 64, LoRA dropout: 0.05, LoRA $\alpha$: 16, and LoRA layers: all layers.

\begin{table*}[ht]
    \centering
    \begin{tabular}{ccccccccc}
    \toprule
         \textbf{Model}&\textbf{Dataset}&\textbf{\makecell{Batch\\Size}} &\textbf{LR}&\textbf{Scheduler}&\textbf{\makecell{Warmup\\Ratio}}&\textbf{Epochs}&\textbf{Max length}&\textbf{\makecell{Zero\\Stage}} \\
    \midrule
       7B &Alpaca&16&2e-4&constant&0.03&3&512&0\\
       7B &LIMA&16&2e-4&constant&0.03&15&4096&2\\
       \midrule
       13B&Alpaca&16&2e-4&constant&0.03&3&512&0\\
       13B&LIMA&16&2e-4&constant&0.03&15&4096&2\\
       \midrule
       70B&Alpaca&32&1e-4&constant&0.03&3&512&2\\
       70B&LIMA&32&1e-4&constant&0.03&15&4096&2(offload)\\
    \bottomrule
    \end{tabular}
    \caption{Training details of SFT method on LIMA and Alpaca Datasets.}
    \label{tab:sft_param}
\end{table*}